\documentclass[letterpaper, 10 pt, conference]{ieeeconf}  

\IEEEoverridecommandlockouts                              

\AtEndDocument{\par\leavevmode} 

\usepackage[draft]{pgf}
\usepackage{physics}
\usepackage{siunitx}
\usepackage{graphics} 
\usepackage{graphicx} 
\usepackage{epsfig} 
\usepackage{amsmath, bm} 
\usepackage{amssymb}  
\usepackage{upgreek}
\usepackage{cite}   

\IEEEtriggeratref{22}

\usepackage{cuted}
\usepackage{flushend}
\usepackage{booktabs}
\usepackage{multicol}
\usepackage{multirow}
\usepackage{stfloats}
\usepackage{lipsum}  
\usepackage{breqn}
\usepackage{comment} 

\usepackage[caption=false]{subfig} 

\graphicspath{ {./images/} }
\title{\LARGE \bf
An Integrated Mechanical Intelligence and Control Approach Towards Flight Control of \textit{Aerobat}
}

\author{Eric Sihite, Atefe Darabi, Pravin Dangol, Andrew Lessieur, and Alireza Ramezani$^{1}$
\thanks{$^{1}$SiliconSynapse Laboratory, ECE Department, Northeastern University, Boston, MA, USA. emails: \{e.sihite, darabi.a, dangol.p, lessieur.a, a.ramezani\} @northeastern.edu}%
}

\begin{document}

\maketitle
\thispagestyle{empty}
\pagestyle{empty}

\begin{abstract}

Our goal in this work is to expand the theory and practice of robot locomotion by addressing critical challenges associated with the robotic biomimicry of bat aerial locomotion. Bats are known for their pronounced, fast wing articulations, e.g., bats can mobilize as many as forty joints during a single wingbeat, with some joints reaching to over one thousand degrees per second in angular speed. Copying bats flight is a significant ordeal, however, very rewarding. Aerial drones with morphing bodies similar to bats can be safer, agile and energy efficient owing to their articulated and soft wings. Current design paradigms have failed to copy bat flight because they assume only closed-loop feedback roles and ignore computational roles carried out by morphology. To respond to the urgency, a design framework called \textit{Morphing via Integrated Mechanical Intelligence and Control (MIMIC)} is proposed. In this paper, using the dynamic model of Northeastern University's \textit{Aerobat}, which is designed to test the effectiveness of the MIMIC framework, it will be shown that computational structures and closed-loop feedback can be successfully used to mimic bats stable flight apparatus. 
 
\end{abstract}


\section{Introduction}
\label{sec:introduction}

In recent years, there has been increased focus on making our residential spaces smarter, safer, more efficient and closer to the materialization of the concept of smart cities \cite{lee2014trends}. As a result, safety and security robots are gaining ever growing importance \cite{pavlidis2001urban} and drive a lucrative market. Smart cities market was valued at USD 624.81 billion in 2019 and is expected to reach USD 1712.83 billion by 2025 \cite{noauthor_smart_nodate}.

Among these security robots, ground robots (wheeled and legged) are used the most despite known limitations such as not being able to reach a vantage point for surveillance, having limited operation time, having the potential to collide with and harm humans in crowded spaces (e.g., sidewalks, airports, etc.), not being able to negotiate rough terrain (e.g., street curbs, stairs, bumps, etc.) or large obstacles (e.g., a wall or bush), and possessing slow mobility \cite{tiddi2020robot,koolen_design_2016,fahmi_passive_2019-2,dario_bellicoso_perception-less_2016,bretl_testing_2008,pardo_evaluating_2016,mastalli_trajectory_2017,mastalli_-line_2015,winkler_gait_2018,aceituno-cabezas_simultaneous_2018,dai_planning_2016}. Despite their superior mobility, which makes them extremely suitable for civic surveillance and monitoring applications, the contribution of Micro Aerial Vehicles (MAVs) to this market has remained very limited. Current state-of-the-art MAVs with fast rotating propellers and rigid structures pose extreme dangers to humans, e.g., they can cause penetrating injury, laceration resulting in blood loss and massive destruction of the human body \cite{arterburn2017faa}. Furthermore, they cannot survive unavoidable crashes in unstructured environments of cities or operate for more than thirty to forty minutes. That said, the application of safety features in these systems have not solved the problems and their operations in residential spaces have remained limited by strict rules from the Federal Aviation Administration (FAA) \cite{arterburn2017faa}. To be able to use MAVs in civic applications, there is a need to transform their safety and efficiency.

\begin{figure}[!t]
    \centering
    \includegraphics[width=1.0\linewidth]{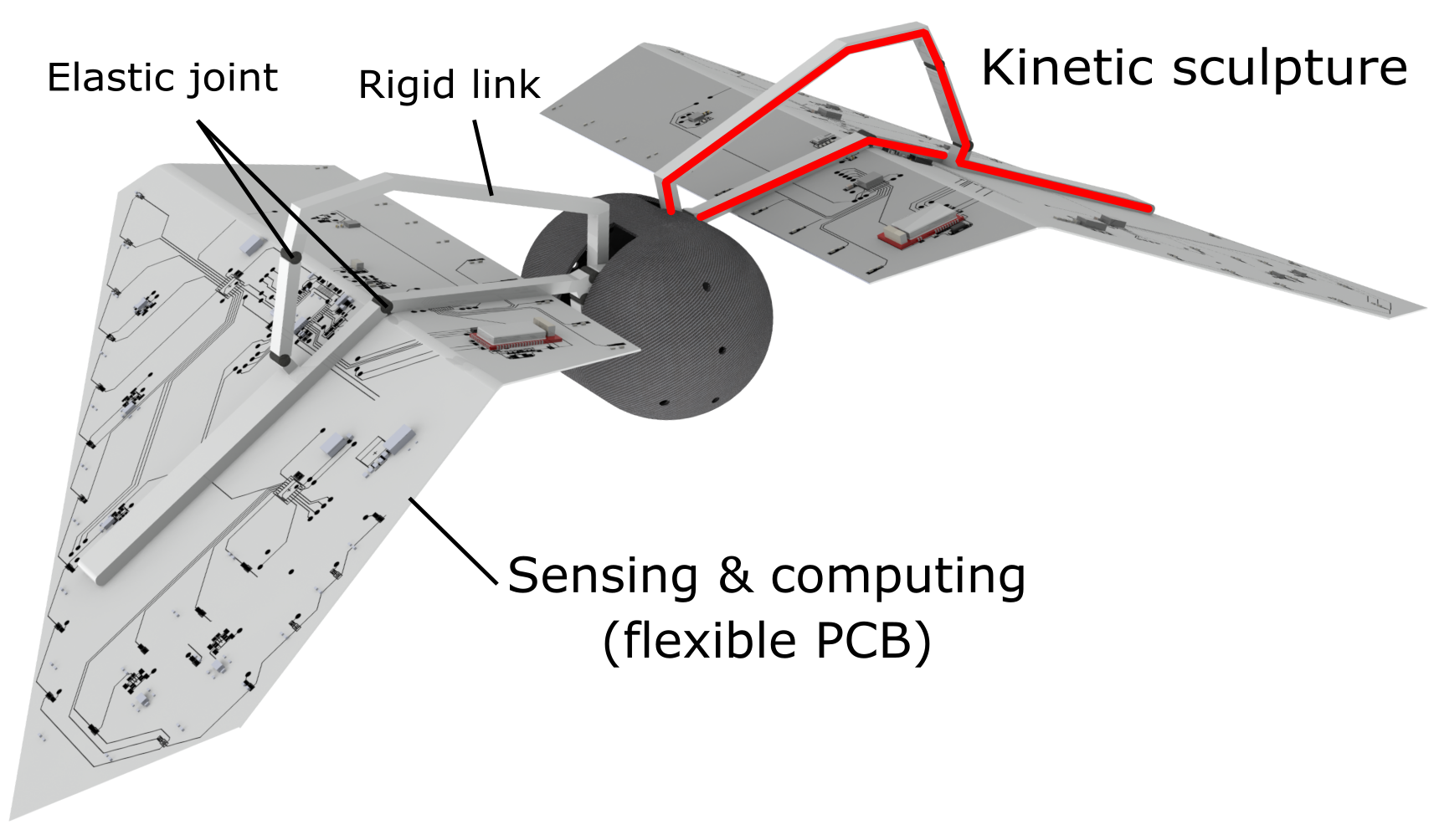}
    \caption{Illustration of Northeastern University's \textit{Aerobat}.}
    \label{fig:cover_image}
\end{figure}
 
\begin{figure*}
    \centering
    \includegraphics[width=0.99\linewidth]{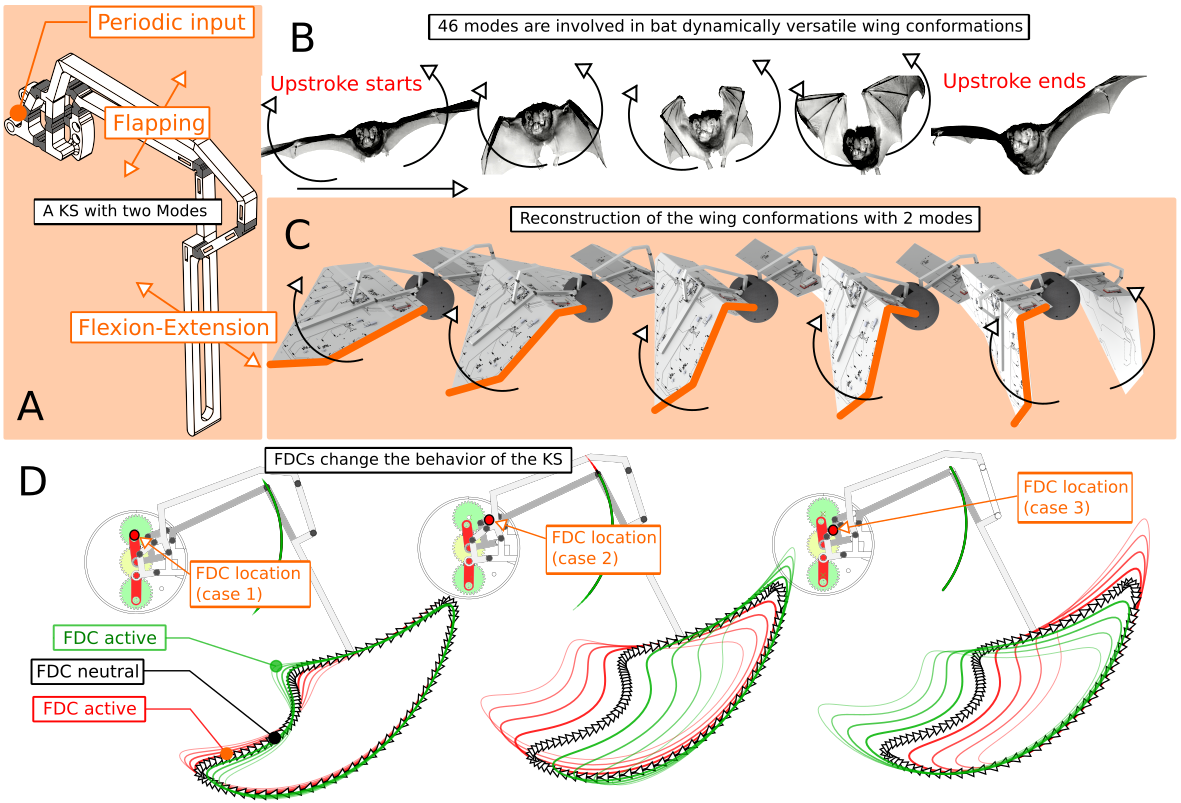}
    \caption{The kinetic sculpture (KS) design and the motivation to mimic a bat's natural flapping gait. (A) Illustrates the KS made of monolithically fabricated rigid and flexible materials. (B) Depicts bat flapping gait with the wing folding/expansion within one flapping cycle. (C) Illustrates simulated Aerobat wingbeat cycle. (D) Shows the sensitivity analysis resutls. Feedback-Driven Components (FDCs) used in the KS can change the behavior of the structure which can be leveraged for flight design purpose.}
    \label{fig:justification}
\end{figure*}

The research goal in this work is to expand the theory and practice of robot locomotion by addressing critical challenges associated with the robotic biomimicry of bat aerial locomotion. The resulting MAVs with morphing bodies will be safe, agile and energy efficient owing to their articulated, soft wings and will autonomously operate with long operation lifespans. Bat membranous wings possess unique functions \cite{tanaka2015flexible} that make them a good example to take inspiration from and transform safety, agility and efficiency of current aerial drones. In contrast with other flying vertebrates, bats have an extremely articulated musculoskeletal system (Fig.~\ref{fig:justification}-B) which is key to their body impact survivability and their impressively adaptive and multimodal locomotion behavior \cite{riskin2008quantifying}. Bats exclusively use this capability with their structural flexibility to generate the controlled force distribution on each membrane wing. Wing flexibility, complex wing kinematics, and fast muscle actuation allow these creatures to change their body configuration within a few tens of milliseconds. These characteristics are crucial to their unrivaled agility and energetic efficiency \cite{azuma2012biokinetics}.

Complex locomotion styles achieved through such synchronous movements of many joints are showcased by several other species. There is an urgency for new paradigms providing insight into how these animals mobilize and regulate so many joints with small brains (limited computation power) and small muscles (limited actuation power). That said, widely used paradigms have failed to copy bat flight \cite{bahlman2013design,colorado2012biomechanics} because they assume only closed-loop feedback roles and ignore computational roles carried out by morphology. To respond to the urgency, a design framework called Morphing via Integrated Mechanical Intelligence and Control (MIMIC) is proposed in this work. In this work, using simulation results, it will be shown that the MIMIC framework can be successfully used to mimic bats flight apparatus known for their pronounced, fast wing articulations, e.g., bats can mobilize as many as forty joints during a single wingbeat, with some joints reaching to over one thousand degrees per second in angular speed. 

We expand upon our previous work with the bat robots \cite{ramezani_biomimetic_2017, hoff_synergistic_2016, hoff_optimizing_2018, hoff2017reducing, ramezani_towards_2020} where we will apply the MIMIC framework to our most recent morphing wing design in \cite{sihite_computational_2020}. This design captures the elbow flexion and extension that allow the wing to fold during upstroke. The MIMIC framework will be the continuation of our past attempts at developing control framework for bio-inspired flapping wing drones in \cite{ramezani_lagrangian_2015, hoff_optimizing_2018, hoff_trajectory_2019,ramezani2016nonlinear,ramezani2017describing,syed2017rousettus,sihite_enforcing_2020}.

This paper is outlined as follows: a brief overview of our bat robot and testing platform, the \textit{Aerobat} as shown in Fig.~\ref{fig:cover_image}, followed by the dynamic modeling, control, optimization, simulation, and concluding remarks.

\section{Overview of Aerobat, a Platform to Test the MIMIC Framework}
\label{sec:mimic_overview}

\begin{figure}[t]
\centering
\includegraphics[width=\linewidth]{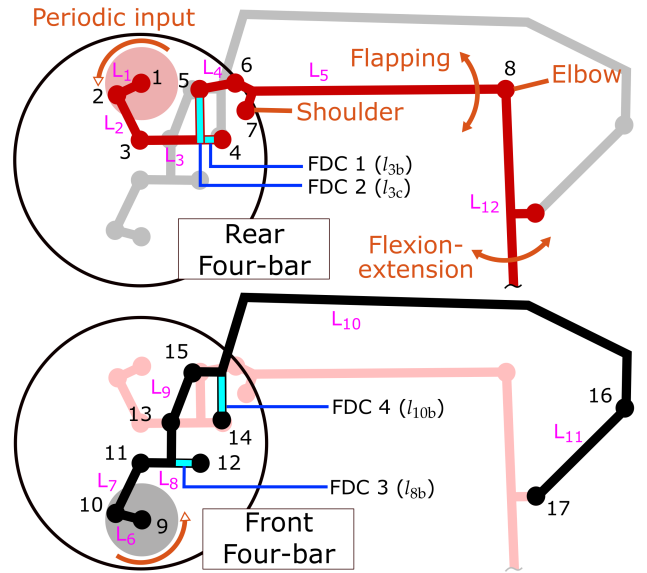}
\caption{Shows the wing structure which is composed of a network of mechanical linkages. The system consists of 12 links ($\{L_1, \dots, L_{12}\}$) and 17 joints/hinges ($\{j_1, ...,j_{17} \}$). Joints 1 and 9 are the gears' center of rotation where they will be driven up to 10 Hz in angular speed. Four FDCs adjust the length of the linkage to change the resulting end-effector trajectories and facilitate control (see the sensitivity analysis results shown above.} 
\label{fig:wing_structure}
\end{figure}

Our robotic bat arm-wing, shown in Fig.~\ref{fig:justification}-A and Fig.~\ref{fig:wing_structure}, is a mechanical structure with computational roles designed to mimic the flying maneuvers of biological bats \cite{sihite_computational_2020}. We also refer to this structure with the term: \textit{Kinetic Sculpture}. The computational roles of this structure are actively modulated using closed-loop feedback as this will be discussed later in this paper. This arm-wing is composed of rigid links and flexible joints monolithically fabricated by using PolyJet 3D printing technology. The mechanism for driving the arm-wing which produces the flapping motion is composed of several gears, cranks, and four-bar mechanisms which are actuated by a single motor. The resulting flapping motion follows some of the biologically meaningful degrees-of-freedom (DoF) present in the natural bat's flapping flight, which in our mechanism are the wing plunging motion and elbow extension/retraction, as shown in Fig.~\ref{fig:justification}-C. The arm-wing extends during the downstroke and retracts during the upstroke which serves to minimize the negative lift and increase the flapping gait efficiency. The dynamic modeling and simulation of Aerobat flapping under four meaningful DoFs (plunging, elbow flexion/extension, mediolateral movement, and feathering) was investigated in \cite{sihite_enforcing_2020} where the wing is modeled as a rigid plate and using optimization to find a steady flapping gait and upside-down perching maneuver.

A more sophisticated structure is proposed to add control and morphological freedom into this design. We propose to use Feedback-Driven Components (FDC) to adjust the length of some links as shown in Fig. \ref{fig:wing_structure}. This serves as a way to change the arm-wing morphology resulting in a change in the end effector trajectories, as shown in Fig.~\ref{fig:justification}-D. The trajectories of interest are the humerus and radius links end effectors, which are shown as $L_{5}$ and $L_{12}$ in Fig.~\ref{fig:wing_structure} respectively. The joint angles of these two links represent the shoulder and elbow angles, which correspond to the plunging motion and the elbow extension respectively. This framework is relevant to the concept of \textit{morphological computation}, where a simple control action can influence and achieve complex manipulation in the body morphology which is commonly seen in nature \cite{hauser2012role}.

In order to simplify the system and facilitate control, the mechanical structure is modeled as a network of massed and massless links connected by compliant joints. Such mechanical systems that include elements with comparatively small or zero mass and inertial distribution are considered as singular mechanical systems. In this system, the massless network is used as a guide for the massed system through linear/torsional spring and damper to simulate the flexible joints of the kinetic sculpture, as shown in Fig. \ref{fig:massed_subsystem}. Here, only the humerus and radius links, in addition to the robot's body, are modeled as a massed system which significantly simplifies the dynamic modeling and simulation.

\section{Dynamic Modeling of Mechanical Computational Structures in Aerobat}
\label{sec:modeling}

This section outlines the derivations of the equations of motion for the massless and massed subsystems as described in Section \ref{sec:mimic_overview}. There are two primary frames of reference considered in this paper: the inertial and body frames. The body frame is defined such that the $x$, $y$, and $z$ axes point towards the robot's front, left and top sides, respectively. The rotation from body frame to inertial frame is defined as $\bm{x} = R_B\,\bm{x}^B$, where the superscript $B$ represents the vectors defined in body frame.

The wing structure is modeled as a planar linkage mechanism and the rotation matrix $R(\theta)$ is used to rotate from link's local frame to the body frame, where $\theta$ is the absolute angle of the link with respect to the $y$-axis of body frame. The wing conformation parameters names and values used to derive the equations in this paper are the same as the wing design in \cite{sihite_computational_2020}, where the length $l_{i}$ is used to describe the length parameters of link $i$ ($L_i$) shown in Fig. \ref{fig:wing_structure}. For links that are not simply straight, the dimensions involved are represented with additional alphabet subscripts (e.g. $l_{3a}$, $l_{3b}$, etc).

\subsection{Massless Subsystem Kinematics Formulation}
\label{subsec:modeling_kinematic}

The equations of motion of the massless subsystem can be modeled from its kinematics. This wing flaps in a planar motion, so for the sake of simplicity, the kinematic equations are derived in 2D coordinates in the body's $y$-$z$ plane. The superscript $B$ for the body frame representation is omitted to simplify the derivations. Also, because of the symmetry of the left and right wings, their kinematics follow the same formulation, therefore a general form of equations are presented in this section. 

As shown in Fig. \ref{fig:wing_structure}, the massless links includes three closed kinematic chains. Additionally, the crank gears are coupled to each other to have the same angular velocity.
Let $\bm p_i$ and $\theta_i$ represent the positions and absolute angles of joint $i \in \{1,\dots,17\}$ about the body frame as labeled in Fig. \ref{fig:wing_structure}. The massless linkage kinematics can be derived from the closed kinematic loop of the four-bar linkages. Derive the positions of the joints 3, 11, and 15 as a function of $\theta_i$ following the kinematic chain from the two closest stationary joints (i.e. joints 1, 4, 9, 12, and 14), represented using the subscripts $A$ and $B$, (e.g. $\bm p_{3A}$). For example, the position of joint 3, $\bm p_{3}$, can be derived as follows
\begin{equation}
\begin{gathered}
     \bm p_{3A} = \bm p_1 + R(\theta_1)[l_1, 0]^\top + R(\theta_2)[l_2, 0]^\top \\
     \bm p_{3B} = \bm p_4 + R(\theta_4)[-l_{3a} - l_{3b}, 0]^\top
\end{gathered}
\end{equation}
where $R(\theta_i)$ is a 2D rotation matrix and $l_j$ are the length variables representing the conformation of link $j$. Then the linkage kinematic constraints can be formed as
\begin{equation*}
    \bm q_1 =[\theta_1,\theta_2, \theta_4,\theta_9,\theta_{10},\theta_{12}, \theta_{13},\theta_{14}, l_{3b}, l_{3c}, l_{8b}, l_{10b}]^\top
\end{equation*}
\begin{equation}
\begin{gathered}
    \bm C(\bm q_1) = 
    \begin{bmatrix}
    \bm p_{3A}-\bm p_{3B}\\
    \bm p_{11A}-\bm p_{11B}\\
    \bm p_{15A}-\bm p_{15B}\\
    \theta_1 - \theta_9 - \Delta \phi
    \end{bmatrix}
    = \bm{0}_{7 \times 1},
\end{gathered}
\label{eq:kinematic_constraints}
\end{equation}
where $\bm q_1$ is formed from the non-constant variables and $\Delta \phi$ is the phase difference between the two crank gears ($\theta_1$ and $\theta_9$). The linkage dimensions are constant except for $l_{3b}$, $l_{3c}$, $l_{8b}$, and $l_{10b}$ which represent the length of the FDCs. 

The equation of motion for the massless subsystem can be derived by taking the second time derivative of $\bm C(\bm q_1)$, which can be rearranged into the following form
\begin{equation}
    \bm {\ddot {C}} = M_{A}(\bm{q}_1) \ddot{\bm{q}}_1+ \bm{h}_{A} (\bm{q}_1,\dot{\bm{q}}_1) = \bm{0}_{7 \times 1}.
\label{eq:kinematic_constraint_eom}
\end{equation}
The input to this subsystem is the acceleration of the crank gear driven by the motor ($\theta_1$) and the FDCs. These accelerations can be formed, using abuse of notation, as follows
\begin{equation}
     \begin{aligned}
     M_{B} &= 
     \begin{bmatrix}
     1 & \bm{0}_{1 \times 11} \\
     \bm{0}_{4 \times 7} & I_{4 \times 4}
     \end{bmatrix}, &
     M_{B} \ddot{\bm{q}}_1 &= \bm{u}_1,
     \end{aligned}
\label{eq:kinematic_inputs}
 \end{equation}
where $\bm{u}_1 = [u_g, u_{3b}, u_{3c}, u_{8b}, u_{10b}]^\top$ represents accelerations, $u_g$ is the acceleration of the crank gear $\theta_1$. Finally, combining \eqref{eq:kinematic_constraint_eom} and \eqref{eq:kinematic_inputs} results in the following equation of motion
\begin{equation}
\begin{gathered}
       M_1 \bm {\ddot q}_1 + \bm h_1 = B_1 \bm u_1 \\ 
       M_1 = \begin{bmatrix}
       M_{A} \\ M_{B}
       \end{bmatrix}, \, \, 
       \bm{h}_1 = \begin{bmatrix}
       \bm{h}_{A} \\ \bm{0}_{5 \times 1}
       \end{bmatrix}, \, \,
       B_1 = \begin{bmatrix}
       \bm{0}_{7 \times 5} \\ I_{5 \times 5}
       \end{bmatrix}
\end{gathered}       
\label{eq:kinematic_eom}
\end{equation}

\subsection{Massed Subsystem Dynamics Formulation}
\label{ssec:modeling_dynamic}

\begin{figure}[t]
\centering
\includegraphics[width=0.8\linewidth]{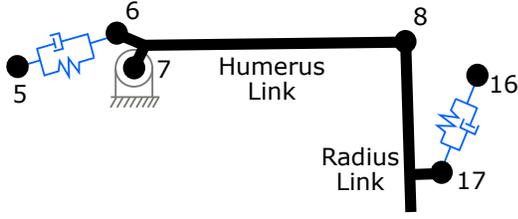}
\caption{Shows the wing massed subsystem which is composed of the humerus and radius links. The external forces and torques imposed by the massless kinematic chain and flexible hinges act on this subsystem.} 
\label{fig:massed_subsystem}
\end{figure}

The equation of motion of the massed subsystem can be derived using the Euler-Lagrangian dynamic formulation. Let $\bm x_{i}$ represent the Center of Mass (CoM) position of the massed subsystem, $i \in \mathcal{L_M} = \{B,H_L,H_R,R_L,R_R\}$, where $B$, $H$, and $R$ represent the body, humerus and radius while the subscripts $L$ and $R$ represent the left and right wing components, respectively. Similar to Section \ref{subsec:modeling_kinematic}, only the general form of the formulations are presented by omitting the $L$ and $R$ subscripts due to the wing symmetry. The positions of the humerus and radius links are defined as follows
\begin{equation}
  \begin{aligned}
\bm x_{H} &= \bm x_{B} + R_B\,  (\bm p_{7}^B + R_x(\theta_{7})\, \bm l_{H}/2 )\\
\bm x_{R} &= \bm x_{H} + R_B\,  (R_x(\theta_{7})\, \bm l_{H}/2 + R_x(\theta_{8})\, \bm l_{R}/2),
\end{aligned}  
\end{equation}
where $R_x$ is the Euler rotation about $x$ axis, $\bm x_B$ is the body inertial position, $\bm l_H = [0,l_h\cos(\alpha),l_{5b} + l_h\sin(\alpha)]^\top$, and $\bm l_R = [0,l_r,0]$. Additionally, $\theta_{7} = \theta_{s} - \alpha$ and $\theta_{8} = \theta_{e}+\theta_{s}+\alpha$, in which $\theta_{s}$ and $\theta_{e}$ are the shoulder and elbow angles of the left wing respectively. $\alpha$, $l_h$, $l_r$, and $l_{5b}$ are the constant physical parameters of the wing structure. Here, $l_h$ and $l_r$ are the length of the humerus and radius links respectively while $\alpha$ corresponds to the initial angle of the shoulder joint at $\theta_7 = 0$.

The generalized coordinates vector is composed of both linear and rotational states, therefore both Hamiltonian and Euler-Lagrangian principles are required to derive the equations of motion of the massed links. In this case, in order to avoid gimbal lock, the body angular velocity is derived by using the modified Euler-Lagrangian equation of motion in SO(3). The kinetic and potential energies of the massed parts and their corresponding Lagrangian, $L = T - U$, can be derived as follows
\begin{equation}
\begin{aligned}
    T &= \textstyle {\frac{1}{2}  \sum_{i \in \mathcal{L_M}} m_{i} {\dot {\bm x}_{i}}^\top {\dot {\bm x}_{i}} + \bm \omega_{i}^{\top} I_{i} \bm \omega_{i}} \\
    U &= \textstyle \sum_{i \in \mathcal{L_M}} m_{i} [0,0,g] \bm x_{i}
\end{aligned}
\end{equation}
where $m_{i}$ is the mass, $I_{i}$ the inertia matrix, and $ \bm \omega_{i}$ the angular velocity in body frame. The angular velocity of every massed link could be computed as the sum of angular velocity of the link and body, $ \bm \omega_{i} = [\dot{\theta}_i,0,0]^\top+\bm \omega_B$ for $i \neq B$.


The equation of motion of the body rotation in SO(3) is defined as follows
\begin{equation}
\begin{aligned}
    R_B^\top = [\boldsymbol{r}_{B1}, \boldsymbol{r}_{B2}, \boldsymbol{r}_{B3} ], \qquad \tfrac{d}{dt} R_B = R_B\,[\bm{\omega}_B]_{\times} \\
    \textstyle{\frac{d}{d t} \left( \frac{\partial L}{\partial \boldsymbol{\omega}_{B}} \right) + \boldsymbol{\omega}_{B} \times \frac{\partial L}{\partial \boldsymbol{\omega}_{B}}+\sum_{j=1}^{3} \boldsymbol{r}_{Bj} \times \frac{\partial L}{\partial \boldsymbol{r}_{Bj}}=\bm u_{2B}}
\end{aligned}
\label{eq:eom_bodyrot}
\end{equation}
where $[\, \cdot \,]_{\times}$ represents the skew symmetric operator and $\bm u_{2B}$ is the generalized force component about the body rotation. Finally, the Euler-Lagrange equations of motion are derived for the rest of the generalized coordinates $\bm q_2 =[\theta_{s_L}, \theta_{e_L}, \theta_{s_R}, \theta_{e_R}, \bm x_B^\top]^\top$, as follows
\begin{equation}
    \textstyle{\dv {}{t} \frac {\partial L} {\partial \dot {\bm q}_2} - {\frac {\partial L} {\partial \bm q_2}} = \bm u_{2A}}.
    \label{eq:eom_therest}
\end{equation}
Then combining \eqref{eq:eom_bodyrot} and \eqref{eq:eom_therest}, the equation of motion of the massed system can be formulated as
\begin{equation}
\begin{gathered}
    M_2 [\ddot{\bm{ q}}_2^\top, \dot{\bm \omega}_B^\top]^\top + \bm h_2 = B_i\bm u_i + B_s\bm u_s + \bm u_a \\
    \dot R_B = R_B\,[\bm{\omega}_B]_{\times},
\end{gathered}    
\label{eq:dynamics_eom}
\end{equation}
where $ M_2$ is the matrix associated with the mass and inertia, and $\bm h_2$ is the vector of gravity and coriolis forces. This system is subjected to several force components, where $\bm{u}_i$ is the internal spring and damping forces of the flexible joints (joints 7 and 8), $\bm{u}_s$ is the actuation forces caused by the massless subsystem, and $\bm{u}_a$ is the generalized aerodynamic force. Finally, combining (\ref {eq:kinematic_eom}) and (\ref {eq:dynamics_eom}) forms the full dynamic model for the numerical simulation.

\subsection{Forces on the Massed System}

\begin{figure}[t]
    \centering
    \includegraphics[width=\linewidth]{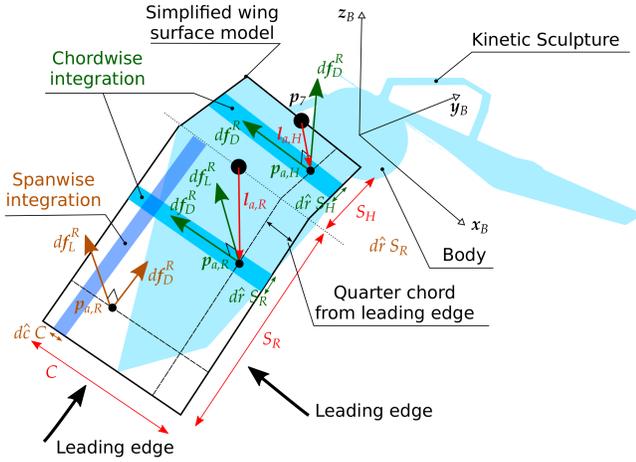}
    \caption{Illustrates the application of aerodynamic strip theory. The overall aerodynamic force is calculated both on spanwise and chordwise strip elements. Then, they are integrated to obtain the resultant aerodynamic force.}
    \label{fig:aerodynamics}
\end{figure}

The flexible joints 7 and 8 are modeled as the combination of a torsional spring and damping which can be derived as $\bm u_i = - ( k_i (\bm \theta_j - \bm \theta_{j,0}) + b_{i}\dot {\bm \theta_j} )$, where $k_{i}$ and $b_{i}$ are the stiffness and damping coefficients respectively, $\bm \theta_j = [\theta_{sL}, \theta_{eL}, \theta_{sR}, \theta_{eR}]^\top$ is the vector containing all massed joint angles, and $\bm \theta_{j,0}$ is the nominal angle of the torsional springs. The $B_i = [I_{4 \times 4}, 0_{4 \times 10}]^\top$ matrix is setup to actuate these joint angles directly. The flexible joints and linkages of the massless subsystem actuates the mass system by the spring and damper model shown in Fig. \ref{fig:massed_subsystem}. Let $\bm u_{s,6}$ and $\bm u_{s,17}$ represent the forces acting on joints 6 and 17, defined as follows
\begin{equation}
\begin{aligned}
    \bm u_{s,6} &= ( k_g\,(|\bm p_5 - \bm p_6| - l_4) + b_g (\dot{\bm p}_5 - \dot{\bm p}_6)^\top \bm e_6 ) \bm e_6 \\
    \bm u_{s,17} &= ( k_g\,(|\bm p_{16} - \bm p_{17}| - l_{11}) + b_g (\dot{\bm p}_{16} - \dot{\bm p}_{17})^\top \bm e_{17} ) \bm e_{17},
\end{aligned}
\label{eq:joint_forces}
\end{equation}
where $\bm e_6 = \frac{\bm p_5 - \bm p_6}{|\bm p_5 - \bm p_6|}$ and $\bm e_{17} = \frac{\bm p_{16} - \bm p_{17}}{|\bm p_{16} - \bm p_{17}|}$ are unit vectors, while $k_g$ and $b_g$ are spring and damping coefficients, respectively. The generalized coordinates can be derived using the virtual displacement and velocity formulation as follows
\begin{equation}
    B_{s,6} = \left( \partial \dot{\bm p}_{6} / \partial \dot{\bm q}_2' \right)^\top, \qquad
    B_{s,17} = \left( \partial \dot{\bm p}_{17} / \partial \dot{\bm q}_2' \right)^\top
\label{eq:joint_forces_B}    
\end{equation}
where $\dot{\bm q}_2' = [\dot{\bm q}_2^\top, \bm \omega_B^\top]^\top$. Then $B_g$ and $\bm u_g$ in \eqref{eq:dynamics_eom} can be formulated as follows 
\begin{equation}
\begin{aligned}
    B_g &= [B_{g,6_L}, B_{g,17_L}, B_{g,6_R}, B_{g,17_R}] \\
    \bm u_g &= [\bm u_{g,6_L}^\top, \bm u_{g,17_L}^\top, \bm u_{g,6_R}^\top, \bm u_{g,17_R}^\top]^\top,
\end{aligned}
\end{equation}
where the subscripts $L$ and $R$ represent the left and right wing respectively. 

In (\ref{eq:dynamics_eom}), $\bm u_a$ represents the generalized aerodynamic force vector acting on the wing composed of lift and drag forces which are generated by the interaction of the wing surface with the airflow. The flapping induces a non-uniform distribution of velocity across the wing surface, therefore the aerodynamic forces must be calculated and then integrated across the wing surface, as shown in Fig. \ref{fig:aerodynamics}. Due to the nature of the flapping motion, the wingtip can also be a leading edge so we also integrate the aerodynamic forces about the wingspan as well. The force is assumed to be concentrated at the aerodynamic center located a quarter chord or wingspan away from the leading edge. 

The lift $d \bm{f}_L$ and drag $d \bm{f}_D$ forces for every segment can be found by the following formulas
\begin{equation}
    \begin{aligned}
      d \bm f^B_L &= \tfrac{1}{2} \rho v_{r}^2 \, C_L(\beta) \, \bm e^B_L \, dS\\
      d \bm f^B_D &= \tfrac{1}{2} \rho v_{r}^2 \, C_D(\beta) \, \bm e^B_D \, dS,
    \end{aligned}
    \label{eq:liftdrag}
\end{equation}
where $\rho$ is the air density, $\bm e_L$ and $\bm e_D$ are the directions of the lift and drag forces, $v_{r}$ is the relative speed of the segment with respect to the wind speed about the axis $\bm e_L$ and $\bm e_D$, $\beta$ is the angle of attack, $C_L$ and $C_D$ are the lift and drag coefficients, and $dS$ the projected wing segment area as shown in Fig. \ref{fig:aerodynamics}. The chordwise integration is defined about the segment surface area $dS = c\,s\,d\hat{r}$ where $c$ is the airfoil chord length, $s$ the wing span length ($s_H$ and $s_R$ for humerus and radius, respectively), and $d \hat r$ the spanwise segment length. On the other hand, the spanwise integration is defined with $dS = c\,s\,d\hat{c}$ where $d\hat{c}$ represents the chordwise segment length. 

The position where the aerodynamic forces are applied, $\bm p_a$, can be found by using the following formulas
\begin{equation}
\begin{aligned}
    \bm p_{a,H}^B &= \bm p_{7}^B + R_x(\theta_{s}) \, \bm{l}_{a,H} \\
    \bm p_{a,R}^B &= \bm p_{7}^B + R_x(\theta_{s})\, \left(
    \bm{l}_{H}^B + R_x(\theta_{e}) \, \bm{l}_{a,R}
    \right),
\end{aligned}
\label{eq:aerodynamic_positions}
\end{equation}
where $\bm{l}_{a,H}$ and $\bm{l}_{a,R}$ are the length vectors from their respective joints to the aerodynamic forces, as illustrated in Fig. \ref{fig:aerodynamics}.  
The relative speed between wing segment and the airflow is computed by
\begin{equation}
    v_{r}^2 = (\bm v_{w}^\top R_B\bm e^B_{L})^2 + (\bm v_{w}^\top R_B\bm e^B_{D})^2,
    \label{eq:windvelocity}
\end{equation}
in which $\bm v_{w} = \bm v_a - \bm v_\infty$ is the relative wind speed, where $\bm v_\infty$ is the true wind speed, and $\bm v_a = \dot{\bm p}_a$ is the inertial velocity of the segment at the location where the aerodynamic force is applied. 

The lift and drag coefficients are defined following the results from \cite{sane2001control} which is experimentally tested for a MAV mimicking a fruit fly. These coefficients are
\begin{equation}
    \begin{aligned}
      C_L(\beta) &= 0.225 + 1.58 \, \sin(2.13 \,\beta - 7.2^\circ) \\
      C_D(\beta) &= 1.92 - 1.55 \, \cos(2.04 \,\beta - 9.82^\circ),
    \end{aligned}
    \label{eq:liftdrag_coeff}
\end{equation}
where $\beta$ is defined in degrees. The angle of attack $\beta$ can be determined by using the following formula
\begin{equation}
    \beta = -\mathrm{atan2}(\bm v_{w}^\top R_B\bm e^B_{L}, \bm v_{w}^\top R_B\bm e^B_{D}).
\end{equation}


Then the generalized aerodynamic force acting on each wing segment can be derived as
\begin{equation}
 d \bm{u}_{a,j} (\hat{x}) = \left( \frac {\partial \dot{\bm{p}}_{a,j} (\hat{x})} {\partial {\dot{\bm q}_2'}} \right)^\top R_B\, 
 \left(
 d \bm{f}^B_{L,j} (\hat{x}) + d \bm{f}^B_{D,j} (\hat{x})
 \right),
\end{equation}
where $\hat{x}$ can either be $\hat{r}$ or $\hat{c}$, and the set $j \in \mathcal{W} = \{H_L, H_R, R_L, R_R\}$ is the set of all wing segments (left/right and humerus/radius). Finally, the resulting generalized aerodynamic forces acting on the system can be solved through the following integration
\begin{equation}
\begin{aligned}
    \bm{u}_a = \sum_{j \in \mathcal{W}} \left( 
    \int_0^1 d \bm{u}_{a,j}(\hat{r})
    \right) + 
    \sum_{j \in \mathcal{W_R}} \left( 
    \int_0^1 d \bm{u}_{a,j}(\hat{c})
    \right),
\end{aligned}
\end{equation}
where the set $\mathcal{W_R} = \{R_L, R_R\}$ is the set of the radius wing segments.

\section{Controller Design, Optimization, and Numerical Simulation}
\label{sec:control_optimization}



The wing mechanism is controlled by adjusting the flapping rate and the length of the FDCs, which are the components of the massless subsystem, through the input $\bm u_1$ in \eqref{eq:kinematic_eom}. In this view, kinetic sculptures deliver dynamic morphing capabilities which are the key features in bats flight apparatus while FDCs take supervisory roles to stabilize the flight dynamics. Let $\dot{\theta}_1$ be the speed of the motor driven crank gear while $\bm{l} = [l_{3b}, l_{3c}, l_{8b}, l_{10b}]^\top$ be the vector containing the length of FDCs. The flapping frequency and the FDC lengths can be adjusted by a simple PD controller as shown below
\begin{equation}
\begin{aligned}
    u_g &= K_{d1} ( \omega_{ref} - \dot{\theta}_1 ) \\
    \bm{u}_p &= K_{p2} \left(\bm l_{ref} - \bm{l} \right) - K_{d2} \, \dot{\bm{l}},
\end{aligned}
\label{eq:controller}    
\end{equation}
where $K_{pi}$ and $K_{di}$ are the control gains, $\omega_{ref}$ is the desired flapping frequency, $\bm{u}_p = [u_{3b}, u_{3c}, u_{8b}, u_{10b}]^\top$, and $\bm l_{ref}$ is the desired FDC length vector. The flapping rate is set to be a constant value of 10 Hz which is the approximate flapping frequency of the Egyptian fruit bat (\textit{rousettus aegyptiacus}) which is the basis of our robot's design. On the other hand, the $\bm l_{ref}$ is found using optimization framework which will be outlined in the following sections.

\subsection{Open Loop Gait Optimization}

\begin{figure}[t]
    \centering
    \includegraphics[width=\linewidth]{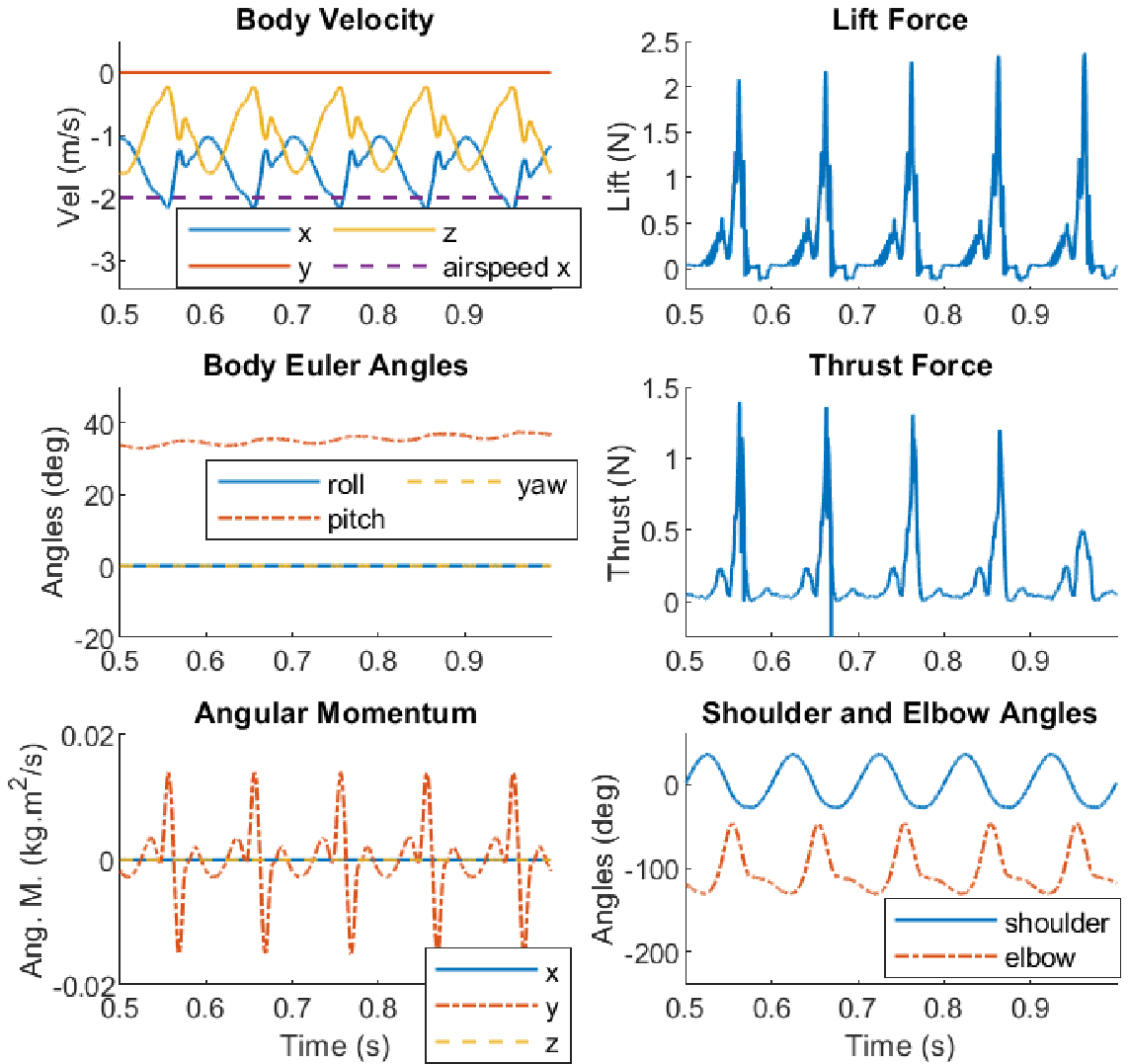}
    \vspace{0.025in}
    
    \includegraphics[width=\linewidth]{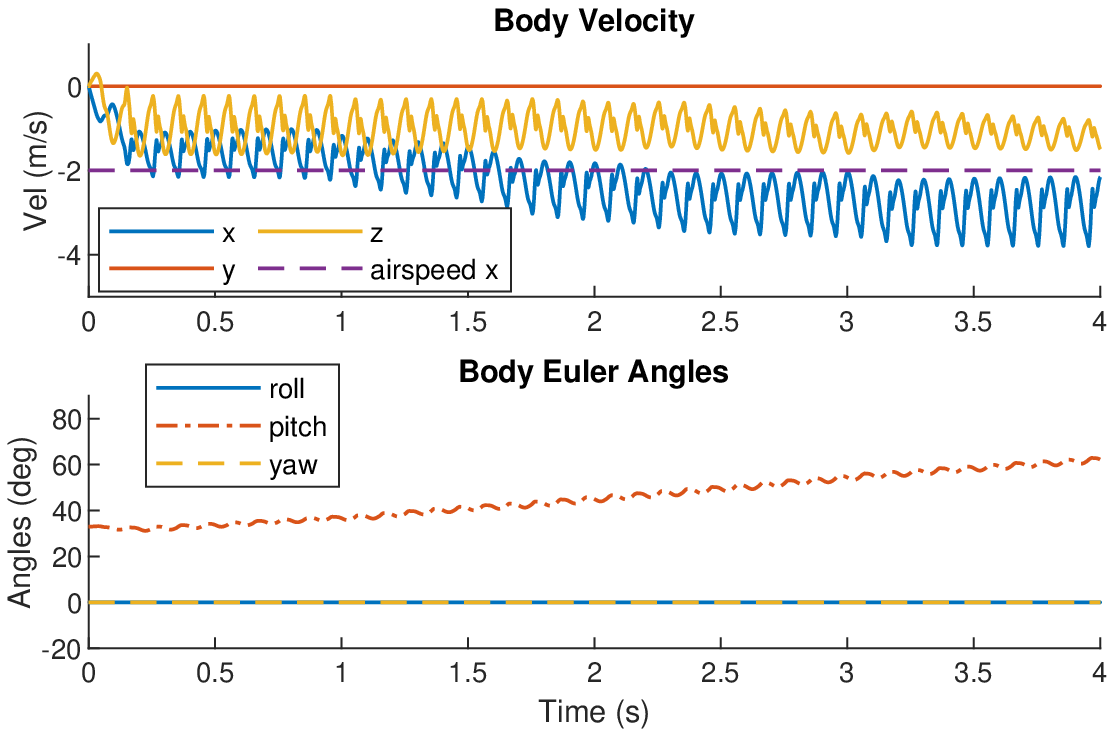}
    \caption{Illustrates the optimized gait. This gait yields stable and periodic trajectories. The gait produces adequate thrust and lift forces. However, as it is seen above, the pitch angle is slowly drifting up.}
    \label{fig:optimization_result_gait}
\end{figure}

This optimization is setup to find a gait which has a stable limit cycle by optimizing the body initial pitch and FDC lengths. The FDC lengths are set to be constant in this optimization, which results in a periodic flapping gait with no feedback stabilization. The feedback into the system will be added after this stable limit cycle is found which will be outlined in the next section. 

The cost function for the optimization is determined from a numerical simulation done over a set period of time. Let the time evolution of the simulated states be $\bm{x}_{k+1} = \bm{f}(\bm{x}_k,\bm{l}_{ref,k}, \bm{u}_k)$, where the simulation states $\bm x_k$ contains all of the relevant states from the equation of motions derived in Section \ref{sec:modeling}. The angular momentum is used as a metric to stabilize the flapping gait which is derived using the following formulations
\begin{equation}
    \bm \Pi = \textstyle \sum_{i \in \mathcal{L_M}} R_B\,I_i\,\omega_i - m_i(\bm x_i - \bm x_{CoM}) \times \dot{\bm x}_i,
\label{eq:ang_momentum}
\end{equation}
where $\bm x_{CoM}$ represents the averaged CoM position of the entire robot. Then the stable limit cycle is found by solving the following optimization problem:
\begin{equation}
    \begin{aligned}
        \min_{\bm{l}_{ref},\theta_y} & \quad J = \sum^N_{k=1} (w_1 \, \bm{\Pi}_k^\top\,\bm{\Pi}_k + w_2 \, \dot{\bm{x}}_{B,k}^\top\,\dot{\bm{x}}_{B,k})\, \Delta t \\
        \mathrm{s. t.}  & \quad \bm{l}_{min} \leq \bm{l}_{ref} \leq \bm{l}_{max},
    \end{aligned}
\label{eq:optimization_gait}
\end{equation}
where the cost $J$ is setup to minimize the angular momentum $\bm{\Pi}$ and body velocity $\dot {\bm{x}}_B$, $\theta_y$ is the initial pitch angle, $\bm{l}_{min}$ and $\bm{l}_{max}$ are the bounds on $\bm{l}_{ref}$, $\Delta t$ is the simulation time step, $N$ is the simulation final step, and $w_i$ is the weight factor in the cost function. The bounds are set at $\bm{l}_{min} = 0.8\,\bm{l}_{0}$ and $\bm{l}_{max} = 1.2\,\bm{l}_{0}$, where $\bm{l}_{0}$ is the initial lengths of FDCs in the wing structure. The robot is initialized at rest and subjected to a wind speed of $\bm{v}_\infty = [-2,0,0]^\top$ m/s. The wing mechanism is set at an offset of -10 cm from the body CoM about the body $x$-axis and the wing chord length is set at 20 cm. The remaining parameters are the same as the kinetic sculpture shown in Fig. \ref{fig:justification}-A and our previous work in \cite{sihite_computational_2020}.

The optimization in \eqref{eq:optimization_gait} is run using an RK4 simulation with a simulation end time of 1 second. The gait optimization result can be seen in Fig. \ref{fig:optimization_result_gait}, where a relatively stable gait is found with the optimal parameters $\bm{l}_{ref} = [7.8, 10.5, 6.2, 7.2]$ mm and $\theta_y = 33^\circ$. The simulation result shows that this gait produces positive lift and thrust while having a periodic and approximately constant velocity of $[-1.52,0,-0.96]$ m/s. A longer simulation time of 4 seconds is also shown in Fig. \ref{fig:optimization_result_gait} where the pitch angle is shown to drifts slowly upwards and the robot is also losing forward speed over time. The drift in the pitch angle will be stabilized by using the FDCs which will be outlined in the next section.

\subsection{Pitch Stabilization Optimization}

\begin{figure}[t]
    \centering
    \includegraphics[width=\linewidth]{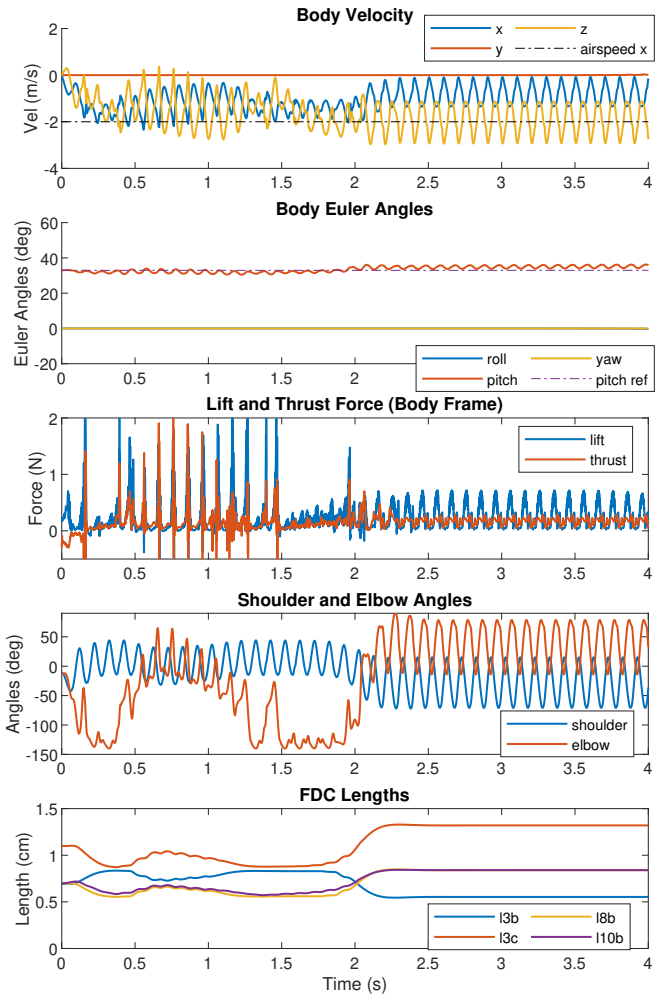}
    \caption{Shows the closed-loop simulation results based on the MIMIC framework, i.e., integrating mechanical intelligence and control. The flight controller has successfully stabilized the pitch angle after a transient period of approximately 2 seconds.}
    \label{fig:simulation_pitch}
\end{figure}

A pitch stabilization controller can then be implemented to regulate the slow pitch drift that exist in the optimized gait. We consider the following controller
\begin{equation}
\begin{aligned}
    \bm l_{ref} = \bm l_{ref,zp} + K_c \, (\theta_{y,ref} - \theta_y),
\end{aligned}
\label{eq:controller_pitch}    
\end{equation}
where the $\bm l_{ref,zp}$ is the constant zero-path flight reference found in the gait optimization \eqref{eq:optimization_gait}, $\theta_{y,ref}$ is the pitch angle reference, and $K_c$ is the controller gain matrix. The gain for the controller in \eqref{eq:controller_pitch} can be found using the following optimization
\begin{equation}
    \begin{aligned}
        \min_{K_c} & \quad \textstyle{ J = \sum^N_{k=1} (w_1 \, \bm{\Pi}_k^\top\,\bm{\Pi}_k + w_2 \, \dot{\bm{x}}_{B,k}^\top\,\dot{\bm{x}}_{B,k} }  \\
        & \qquad \qquad \qquad + w_3 \, (\theta_{y,ref} - \theta_{y_k})^2)\, \Delta t \\
        \mathrm{s. t.}  & \quad K_{c,min} \leq K_{c} \leq K_{c,max}.
    \end{aligned}
\label{eq:optimization_pitch}
\end{equation}
This optimization has an additional cost $w_3$ for weighting the robot's pitch versus a constant pitch reference. The FDC lengths are also constrained using the saturation $\bm{l}_{min} \leq \bm{l}_{ref} \leq \bm{l}_{max}$ to prevent the lengths from going unbounded.

%
%

The optimization found an optimal controller gain of $K_c = [0.42, -0.26, -0.38, -0.097]^{\top}$ and the simulation result is shown in Fig. \ref{fig:simulation_pitch} which has a transient state up to $t = 2$s before reaching a steady limit cycle. The pitch is stable near the target pitch angle of $33^{\circ}$ throughout the simulation. This shows that the controller has successfully achieved pitch stabilization by utilizing the FDCs.


\section{Conclusions and Future Work}
\label{sec:conclusions}

Copying bat high-dimensional flight apparatus is nearly impossible by considering classical feedback design paradigms based on sensing, computing and actuation. Bat musculoskeletal system possesses many active and passive joints, e.g., bat wings possess over 40 joints. In this work, we offered a solution towards the robotic biomimicry of bat aerial locomotion. We proposed a design framework called Morphing via Integrated Mechanical Intelligence and Control (MIMIC). We leveraged computational structures called Kinetic Sculptures (KS) to subsume part of the responsibility of closed-loop feedback under mechanical intelligence. Then, we extended our previous works by considering Feedback-Driven Components (FDC) in the design of KS which possess supervisory roles and are used for flight stabilization. We used the dynamical model of Northeastern University's Aerobat to show the successful stabilization of the Aerobat's longitudinal dynamics. This robot possesses an articulated wing structure with many active and passive joints. In addition, the robot is tail-less which means its longitudinal dynamics are open-loop unstable.



\bibliographystyle{IEEEtran}
\bibliography{ref, references-eric}








\end{document}